\documentclass[twocolumn, 12pt]{article}

\usepackage{titling}
\title{\vspace{-1.8cm}\bfseries{\Large{Continual Weight Updates and Convolutional Architectures for Equilibrium Propagation}} \\\vspace{-0.5cm}}

\author{\normalsize {Maxence Ernoult\thanks{Centre de Nanosciences et de Nanotechnologies, Universit\'e Paris-Saclay}\textsuperscript{      , 2}, Julie Grollier \thanks{Unit\'e Mixte de Physique, CNRS, Thales, Universit\'e Paris-Saclay}, Damien Querlioz\textsuperscript{1}, Yoshua Bengio\thanks{Canadian Institute for Advanced Research}\textsuperscript{      , 4}, Benjamin Scellier\thanks{Mila, Universit\'e de Montr\'eal}}}

\thanksmarkseries{arabic}

\date{}
\usepackage[T1]{fontenc}
\usepackage[utf8]{inputenc} 
\usepackage[english]{babel}
\usepackage{amssymb,amsthm,amsfonts,amscd}
\usepackage{amsmath}
\usepackage{mathtools}   
\usepackage{fancyhdr}    
\usepackage{makeidx} 
\usepackage{graphicx}
\usepackage{algorithm}
\usepackage{algorithmic}
\usepackage{array}
\usepackage{caption}
\usepackage{lipsum}
\graphicspath{{.},{img/}}

\makeindex
\usepackage{geometry}
\newcommand{\norm}[1]{\left\lVert #1\right \rVert}  

\begin{document}
\maketitle
\subsection*{Summary}
Equilibrium Propagation (EP) \cite{EP} is a biologically inspired alternative algorithm to backpropagation (BP) for training neural networks.
It applies to RNNs fed by a static input $x$ that settle to a steady state, such as Hopfield networks.
EP is similar to BP  in that in the second phase of training, an error signal propagates backwards in the layers of the network, but contrary to BP, the learning rule of EP is spatially local.
Nonetheless, EP suffers from two major limitations. 
On the one hand, due to its formulation in terms of real-time dynamics, EP entails long simulation times, which limits its applicability to practical tasks.
On the other hand, the biological plausibility of EP is limited by the fact that its learning rule is not local in time:
the synapse update is performed after the dynamics of the second phase have converged and requires information of the first phase that is no longer available physically.
Our work addresses these two issues and
aims at widening the spectrum of EP from standard machine learning models to more bio-realistic neural networks.
First, we propose a discrete-time formulation of EP which enables to simplify equations, speed up training and extend EP to CNNs. Our CNN model achieves the best performance ever reported on MNIST with EP. 
Using the same discrete-time formulation, we introduce \emph{Continual Equilibrium Propagation} (C-EP): 
the weights of the network are adjusted continually in the second phase of training using local information in space and time. We show that in the limit of slow changes of synaptic strengths and small nudging, C-EP is equivalent to BPTT (Theorem~1). We numerically demonstrate Theorem~1 and C-EP training on MNIST and generalize it to the bio-realistic situation of a neural network with asymmetric connections between neurons.

\vspace{-0.5cm}
\subsection*{A simplified formalism}

In the original formulation of EP \cite{EP}, the dynamics of the neurons (denoted $s_t$) is in real time and derives from an energy function: $\frac{ds_t}{dt}=-\frac{\partial E}{\partial s}(s_t)$. 
By discretizing the real-time dynamics with a time discretization parameter $\epsilon$ and defining $\Phi = \frac{1}{2}\norm{s}^2 - \epsilon E(s)$, the dynamics rewrites:
\begin{equation}
    s_{t+1} = \frac{\partial \Phi}{\partial s}(x,s_t,\theta),
    \label{eq:formalism}
\end{equation}
Our first contribution is to generalise EP to any $\Phi$.

\paragraph{Algorithm.}
EP proceeds in two phases.
In the first phase, neurons evolve freely following Eq.~(\ref{eq:formalism}) towards a first steady state $s_*$.
In the second phase, the subset of output neurons $\hat{y}$ is elastically nudged towards a target $y$ until the neurons reach a new steady state $s_*^\beta$.
More technically $s_{t+1}^\beta = \frac{\partial \Phi}{\partial s}(x,s_t^\beta,\theta) - \beta\frac{\partial\ell}{\partial s}(s_t^\beta,y)$  with $\ell(s,y) = \frac{1}{2}\norm{\hat{y} - y}$ being the cost function.
The learning rule reads:

\begin{equation}
    \Delta\theta = \frac{1}{\beta}\left(\frac{\partial\Phi}{\partial\theta}(x,s_*^\beta,\theta) - \frac{\partial\Phi}{\partial\theta}(x,s_*,\theta)\right)
    \label{eq:learning-rule}
\end{equation}

\subsection*{Towards conventional machine learning: discrete-time CNNs}
We define our CNN model through the dynamics:
\begin{equation}
    s_{t+1} = \sigma(\mathcal{P}(\theta \star s_{t})),
    \label{eq:cnn-eq}
\end{equation}
where $\sigma$, $\star$ and $\mathcal{P}$ respectively denote an activation function, convolution and pooling.
We show that there exists a primitive function $\Phi$ such that $s_{t+1}\approx\frac{\partial \Phi}{\partial s}(s_t, \theta)$ which, using Eq.~(\ref{eq:learning-rule}), enables to derive:
\begin{equation}
\Delta\theta = \frac{1}{\beta}\left(\mathcal{P}^{-1}(s^\beta_*) \star s^\beta_* - \mathcal{P}^{-1}(s_*) \star s_*\right).
\end{equation}
Our CNN model achieves $\sim 1 \%$ test error on MNIST. More generally, training fully connected architectures with equations of the kind of Eq.~(\ref{eq:cnn-eq}) yields an acceleration of a factor 5 to 8 without loss of accuracy compared to standard EP and BPTT.

\subsection*{Towards more biological plausibility: continual weight updates}

We define \emph{Continual Equilibrium Propagation} as a variant of EP where the first phase still reads as Eq.~(\ref{eq:formalism}) while during the second phase, both neurons and synapses evolve as dynamical variables:
\begin{align}
    \left\{
\begin{array}{ll}
   \displaystyle s_{t+1}^{\beta, \eta} &= \frac{\partial \Phi}{\partial s}(s_{t}^{\eta, \beta}) - \beta\frac{\partial \ell}{\partial s}(s_{t}^{\eta, \beta})\\
    \displaystyle \theta_{t+1}^{\eta, \beta} &= \theta_{t+1}^{\eta, \beta} + \frac{\eta}{\beta}\left( \frac{\partial \Phi}{\partial \theta}(s_{t+1}^{\eta,\beta}) - \frac{\partial \Phi}{\partial \theta}(s_{t}^{\eta,\beta})\right).
 \displaystyle 
\end{array}
\right.
\label{eq:cep-dynamics}
\end{align}
We validate C-EP with training experiments on MNIST, whose accuracy approaches the one obtained with standard EP.

\vspace{-0.6cm}
\subsection*{Equivalence of EP and BPTT}
The BPTT approach for training is top-down: 
defining the loss $\mathcal{L} = \ell(s_*,y)$,
it computes the gradients
\vspace{-0.2cm}
\begin{equation}
   \nabla_{s}^{\rm BPTT}(t) = \frac{\partial \mathcal{L}}{\partial s_{T-t}}, \qquad
   \nabla_{\theta}^{\rm BPTT}(t) = \frac{\partial \mathcal{L}}{\partial \theta_{T-t}},
\label{eq:ep-bptt}
\end{equation}
which in turn determine the weight updates.
In contrast, our approach is bottom-up: we first define the dynamics of Eq.~(\ref{eq:cep-dynamics}), then the EP updates
\vspace{-0.2cm}
\begin{align}
    \left\{
\begin{array}{ll}
   \Delta_{s}^{\rm EP}(\eta, \beta, t) & =\frac{1}{\beta}(s_{t+1}^{\eta,\beta} - s_{t}^{\eta,\beta}),\\
   \Delta_{\theta}^{\rm EP}(\eta, \beta, t) & =\frac{1}{\beta}(\frac{\partial \Phi}{\partial\theta}(s_{t+1}^{\eta, \beta}) - \frac{\partial \Phi}{\partial\theta}(s_{t+1}^{\eta, \beta})),
\end{array}
\right.
\label{eq:ep-bptt}
\end{align}
which we show to compute the gradients of the loss.
Theorem~1 states that, provided the first steady state is maintained long enough, the updates of EP are equal to the gradients obtained by BPTT:
\vspace{-0.3cm}
\begin{align}
    \left\{
\begin{array}{l}
   \displaystyle \lim_{\eta \to 0} \lim_{\beta \to 0} \Delta_{s}^{\rm EP}(\eta, \beta,t)  = -\nabla_{s}^{\rm BPTT}(t),\\
    \displaystyle \lim_{\eta \to 0} \lim_{\beta \to 0} \Delta_{\theta}^{\rm EP}(\eta, \beta,t)  = -\nabla_{\theta}^{\rm BPTT}(t)
 \displaystyle.
\end{array}
\right.
\label{eq:ep-bptt}
\end{align}
Theorem~1 is illustrated on Fig.~\ref{fig}~(a) with a computational graph, where the updates of EP on the right match in the same colour the gradients provided by BPTT on the left. Fig.~\ref{fig}~(b) numerically demonstrates Theorem~1 for a system which exactly fulfills the conditions stated above with $\eta = 0$ ('Standard EP') and $\eta > 0$ ('Continual EP'): plain and dashed lines (i.e. $\Delta_{\theta}^{\rm EP}$ and $-\nabla_{\theta}^{\rm BPTT}$ processes) perfectly coincide and split apart upon using a finite learning rate.
Note that the property also holds even in the non-ideal setting of our CNN model ('Discrete EP').

\vspace{-0.4cm}
\subsection*{Continual weight updates and asymmetric connections between neurons}
\begin{figure}[t]
\begin{center}
 \includegraphics[width=0.4\textwidth]{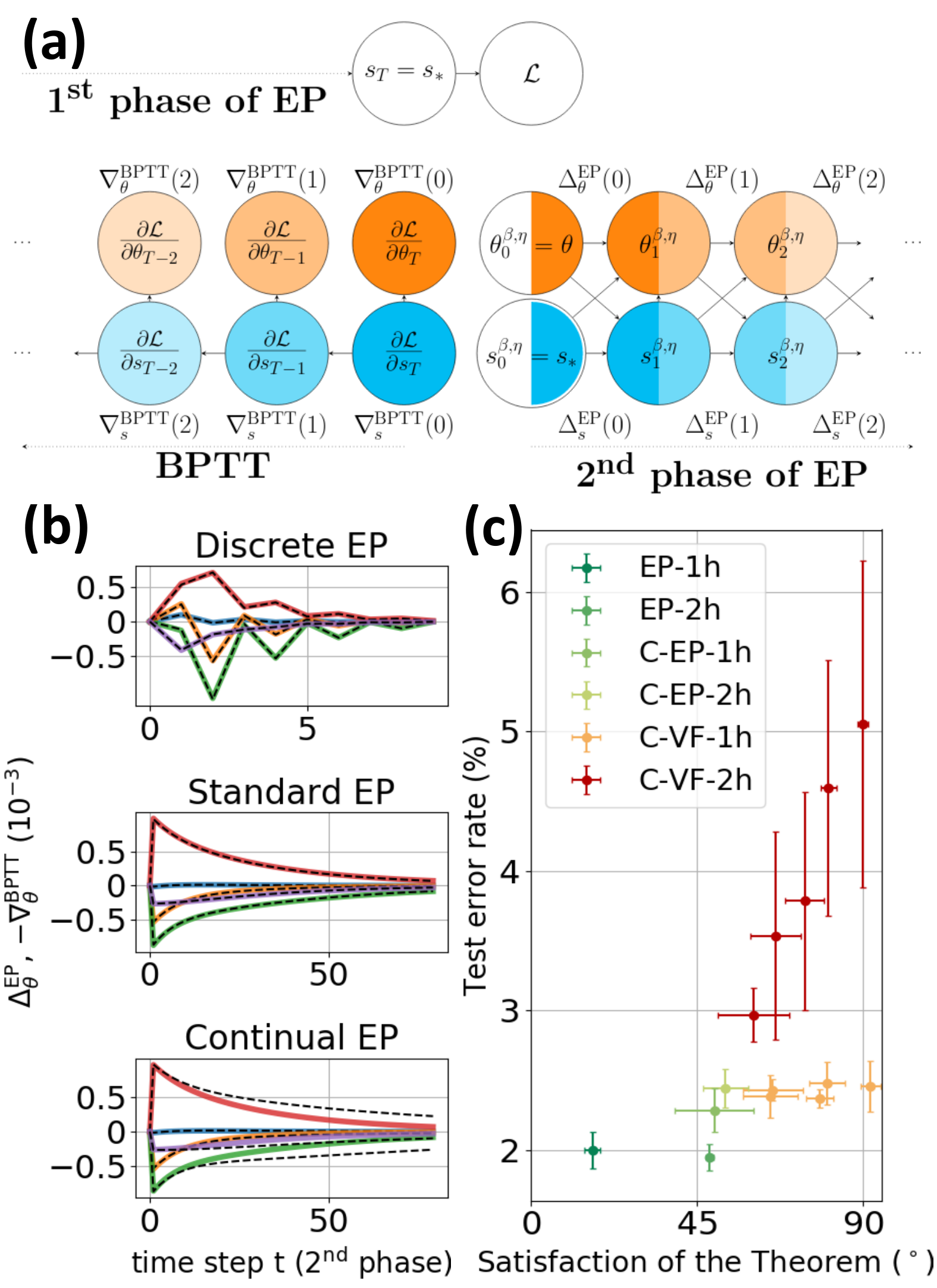}
\end{center}
  \caption{(a) Equivalence of EP and BPTT in the general case where updates are continual ($\eta >0$). (b) Numerical validation of Theorem~1. Plain and dashed line represent $\Delta^{\rm EP}_{\theta}$ and $-\nabla^{\rm BPTT}_{\theta}$ processes respectively. Each color stands for a synapse. (c) Error rate on MNIST as a function of the degree of satisfaction of Theorem 1. Different C-VF points correspond to different weight angles between forward and backward weights, before training.
  }
  \label{fig}
\end{figure}

We extend C-EP to networks whose synaptic connections are asymmetric \cite{VF}, which we name \emph{Continual Vector Field EP} (C-VF). 
On top of demonstrating learning with C-VF on MNIST, we further show numerically on Fig~\ref{fig}~(c) that
the more a network satisfies Theorem~1 before training, the best it can learn.

\vspace{-0.5cm}
\begin{thebibliography}{5}
\bibitem{EP}Scellier, B., \& Bengio, Y. (2017), Frontiers in computational neuroscience, 11, 24.
\bibitem{VF}Scellier, B., Goyal, A., Binas, J., Mesnard, T., \& Bengio, Y. (2018), arXiv preprint arXiv:1808.04873.
\end{thebibliography}
\end{document}